\documentclass[conference]{IEEEtran}
\IEEEoverridecommandlockouts
\usepackage{comment}
\usepackage{amsmath,amssymb,amsfonts}
\usepackage{algorithmic}
\usepackage{graphicx}
\usepackage{textcomp}
\usepackage{wrapfig}
\usepackage{xcolor}
\usepackage{graphicx}
\usepackage{amsmath}
\usepackage{cite}
\usepackage{wrapfig}
\usepackage{soul}
\usepackage{listings}
\usepackage{tikz}
\usepackage[switch]{lineno}
\usepackage{pgfplots, pgfplotstable}
\usepgfplotslibrary{colormaps}%
\pgfplotsset{compat=1.15}
\pgfplotsset{
    colormap={slategraywhite}{
        rgb255=(112,128,144)
        rgb255=(255,159,101)
    }}

\usepackage[colorlinks,urlcolor=blue,linkcolor=blue,citecolor=blue]{hyperref}

\usepackage{mdframed}
\usepackage{tcolorbox}
\def\BibTeX{{\rm B\kern-.05em{\sc i\kern-.025em b}\kern-.08em
    T\kern-.1667em\lower.7ex\hbox{E}\kern-.125emX}}
\begin{document}

\title{\texttt{Cook-Gen}: Robust Generative Modeling of Cooking Actions from Recipes}

\makeatletter
\newcommand{\linebreakand}{%
  \end{@IEEEauthorhalign}
  \hfill\mbox{}\par
  \mbox{}\hfill\begin{@IEEEauthorhalign}
}
\makeatother

\author{\and
\IEEEauthorblockN{\textbf{Revathy Venkataramanan}}
\IEEEauthorblockA{Artificial Intelligence Institute\\
University of South Carolina\\
Columbia, SC, USA \\
\texttt{revathy@email.sc.edu}}
\and
\IEEEauthorblockN{\textbf{Kaushik Roy}}
\IEEEauthorblockA{Artificial Intelligence Institute\\
University of South Carolina\\
Columbia, SC, USA \\
\texttt{kaushikr@email.sc.edu}}
\and
\IEEEauthorblockN{\textbf{Kanak Raj}}
\IEEEauthorblockA{Artificial Intelligence Institute\\
University of South Carolina\\
Columbia, SC, USA \\
\texttt{kanakr@mailbox.sc.edu}}
\linebreakand
\IEEEauthorblockN{\textbf{Renjith Prasad}}
\IEEEauthorblockA{Artificial Intelligence Institute\\
University of South Carolina\\
Columbia, SC, USA \\
\texttt{kaippilr@mailbox.sc.edu}}
\and
\IEEEauthorblockN{\textbf{Yuxin Zi}}
\IEEEauthorblockA{Artificial Intelligence Institute\\
University of South Carolina\\
Columbia, SC, USA \\
\texttt{yzi@email.sc.edu}}
\and
\IEEEauthorblockN{\textbf{Vignesh Narayanan}}
\IEEEauthorblockA{Artificial Intelligence Institute\\
University of South Carolina\\
Columbia, SC, USA \\
\texttt{vignar@sc.edu}}
\linebreakand
\IEEEauthorblockN{\textbf{Amit Sheth}}
\IEEEauthorblockA{Artificial Intelligence Institute\\
University of South Carolina\\
Columbia, SC, USA \\
\texttt{amit@sc.edu}}
}
\maketitle
\begin{abstract}
As people become more aware of their food choices, food computation models have become increasingly popular in assisting people in maintaining healthy eating habits. For example, food recommendation systems analyze recipe instructions to assess nutritional contents and provide recipe recommendations. The recent and remarkable successes of generative AI methods, such as auto-regressive large language models, can lead to robust methods for a more comprehensive understanding of recipes for healthy food recommendations beyond surface-level nutrition content assessments. In this study, we explore the use of generative AI methods to extend current food computation models, primarily involving the analysis of nutrition and ingredients, to also incorporate cooking actions (e.g., add salt, fry the meat, boil the vegetables, etc.). Cooking actions are notoriously hard to model using statistical learning methods due to \textit{irregular data patterns} - significantly varying natural language descriptions for the same action (e.g., marinate the meat vs. marinate the meat and leave overnight) and infrequently occurring patterns (e.g., add salt occurs far more frequently than marinating the meat). The prototypical approach to handling irregular data patterns is to increase the volume of data that the model ingests by orders of magnitude. Unfortunately, in the cooking domain, these problems are further compounded with larger data volumes presenting a unique challenge that is not easily handled by simply scaling up. In this work, we propose novel aggregation-based generative AI methods, \texttt{Cook-Gen}, that reliably generate cooking actions from recipes, despite difficulties with irregular data patterns, while also outperforming Large Language Models and other strong baselines. 
\end{abstract}

\section{Introduction}\label{sec:intro}
Alice Waters once said, "We are what we eat, " succinctly summarizing the impact that food can have on people's physical and mental well-being. Increased access to knowledge about the impact of food choices, along with rapid advances in health monitoring technology, has led to individuals being more aware and cautious about their dietary choices \cite{stehr2020multi}. There has been a growing trend to utilize AI-based diet management systems that analyze food recipes to provide an enhanced food recommendation experience to the user \cite{jospe2015diet,wang2016diet}. Most of these systems incorporate discriminative models for the identification of nutritional information in recipes to encourage healthy eating habits (e.g., ingredients and their nutritional values). However, these models lack a more comprehensive understanding of the recipe. For example, they often do not account for cooking actions when providing recommendations \cite{toledo2019food}. It is vital to include cooking actions in the decision-making of diet management systems since they can significantly impact the nutritional value of food\cite{retinoids2006}, ensure food safety\cite{mccabe2004food}, and even produce harmful substances. For instance, grilling meat at high temperatures produces cancer-causing compounds\cite{farhadian2011effects, dascomprehensive}. Figure \ref{fig:cooking_process_aware_AI} illustrates a recommendation system that analyses the effect of cooking actions and its impact on the final recommendation.
\begin{figure}[!htb]
    \centering
    \includegraphics[width=\linewidth]{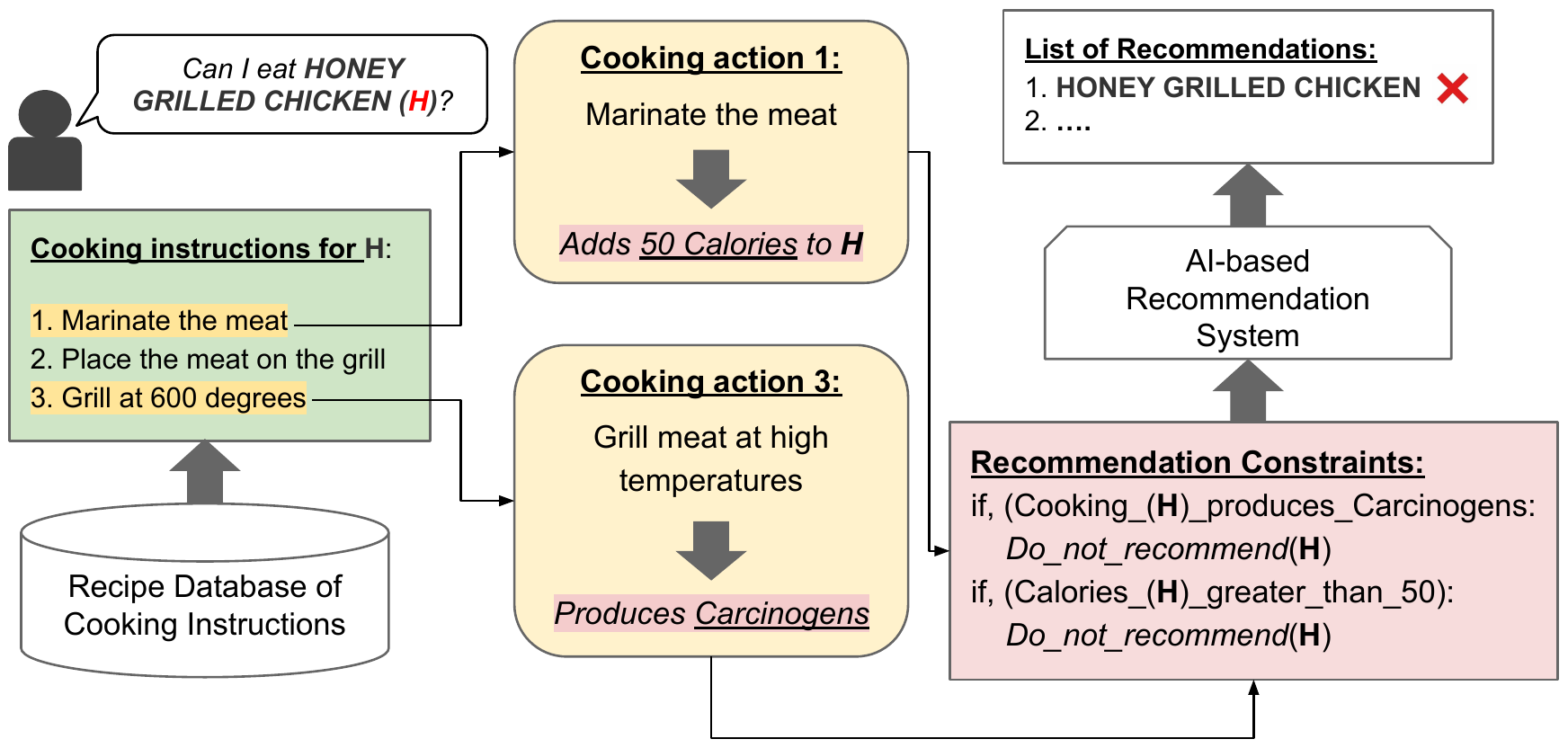}
    \caption{The figure illustrates how understanding and modeling the cooking actions, not just the ingredients and their nutrition values, significantly impacts the final recommendation. Here meat inherently has nothing unhealthy about it. However, marinating it and grilling at high temperatures causes changes to its ``healthiness'' value.}
    \label{fig:cooking_process_aware_AI}
\end{figure}
Recent successes in Large Language Models (LLMs) have the potential to enable a more comprehensive analysis of recipes, such as calorie count, macro-nutrient content, and ingredient lists. This information can then be used to identify healthy recipes and also suggest ingredient substitutions as per the user's health condition and food preferences \cite{shirai2021identifying}. LLMs can further be used to identify common cooking actions, ingredients, and flavor profiles associated with healthy dishes, which can be used to develop healthy eating recommendations and meal plans. 

However, recipes written in natural language pose unique challenges with regard to natural language processing (NLP) that are not observed in traditional NLP tasks. For example, recipes include highly variable textual descriptions for the same cooking action (some people write `` marinate the chicken and leave overnight'' whereas others simply write ``marinate the chicken''). Furthermore, recipes consist of several elements such as ingredients, utensils, temperature, and cooking actions (e.g., ``marinate 1/2 cup of sliced chicken in a bowl'') and, crucially, the frequency of cooking actions (``marinate'') relative to ingredients, quantities, and other information (``1/2, ``bowl and chicken'') is low. We refer to the issues of low-frequency words and variation in lengths of cooking action descriptions collectively as \textit{irregular data patterns}. Thus, discriminative models struggle due to the lack of distributional patterns with a high discriminatory value for expressing complex relationships between different variables (e.g., the effect of cooking actions). 

With the rise of LLMs that have demonstrated remarkable successes with neural network-enabled generative models, AI systems can potentially model more complex phenomena, such as the impacts of cooking actions from the natural language text in recipes \cite{aydin2023chatgpt}. However, these models still make a discriminative model-type decision repeatedly (e.g., continuous next-word prediction), except that they do it using large neural networks accompanied by enormous volumes of training data. Therefore, the problems with \textit{irregular data patterns} for discriminative modeling persist and are, in fact, further compounded at larger scales (e.g., imbalanced frequencies get more imbalanced). Inspired by recent innovations in dynamical systems control that utilize aggregation methods to handle \textit{irregular data patterns} by (1) distilling more regular aggregated information from irregular data patterns and (2) leveraging high capacity function approximation capabilities of higher-order polynomial fits to model complex relationships, we propose \texttt{Cook-Gen}, a novel generative-AI method to generate cooking actions from recipe texts \cite{yu2023moment}. We compare \texttt{Cook-Gen} with state-of-the-art generative and discriminative baseline models and find that it consistently outperforms these baselines.

\section{Related Work}
Here we review previous work on modeling cooking actions from recipes.
Papadopoulos et al. represent cooking recipes using manually specified program abstractions which included cooking actions, quantities, ingredients, and other information from the recipes \cite{papadopoulos2022learning}. Recipescape, a cooking analysis tool, used Stanford's CoreNLP POS Tagger to extract verbs from recipes as a proxy for cooking actions \cite{chang2018recipescape}. Walter et al. also used a POS tagger in a similar context to reconstruct recipes using a workflow diagram of cooking actions. Shidochi et al. manually extracted cooking actions from 15 recipes to suggest substitute actions with similar effects \cite{shidochi2009finding}. Silva et al. used an off-the-shelf annotation tool to annotate recipes for cooking actions as part of an information retrieval system \cite{silva2019information}. A common theme among the existing works is the manual effort involved in cooking action modeling, underscoring the challenge of automated modeling of cooking actions from recipes.

\section{Description of the Dataset and Task, Training and Evaluation Metrics}
\subsection{Dataset Construction and Description}\label{subsec:data}
We utilize the Recipe1M dataset, which is a comprehensive collection of 1 million high-quality recipes containing titles, ingredients, cooking instructions, and recipe images for each entry \cite{salvador2017learning}. We then asked graduate students between ages 23 and 27 from the University of South Carolina to identify cooking actions in the recipe data for a randomly selected sample of 10000 recipe texts. Examples of recipe texts include, ``bring the water to boil'', ``saute the vegetables in the wok'', or `` marinate the chicken''. Their corresponding cooking actions are, ``boil'', ``saute'' and ``marinate''.  

\subsection{Task Description}
As mentioned in Section \ref{sec:intro}, we experiment with both discriminative and generative baselines. We describe the task description in the discriminative and generative modeling settings.
\subsubsection*{\textbf{Discriminative Modeling Setting}} \label{subsubsec:DM}
Let $R$ be the set of all recipes. Each recipe $r \in R$ is written in natural language using a list of words $W_r$ and also consists of a title $T_r$, a list of ingredients $I_r$, and a list of cooking actions $A_r$. Let $S = \{w_m \mid m \gets \{1,\hdots,M\}\}$ denote the set of all $M$ words across the $N$ recipes. Thus we have a recipe set defined as: 
\begin{equation}\label{eqn:recipes}
R = \{(W_r,t_r,l_r,A_r) \mid W_r \subseteq S, t_r \subset S, I_r \subset S, A_r \subset S\}
\end{equation}

We define a function $f: S \rightarrow \{0,1\}$ that predicts whether a given word $w_m \in S$ is a cooking action or not. Thus we define the task of predicting cooking actions in a recipe as a binary classification task, where for each recipe $r$, we want to predict a sequence of binary labels $y_r = \{y_1, \hdots, y_M\}$, where $y_m = (f(w_m) \land \mathbb{I}(w_m \in W_r))$. Here $\mathbb{I}(w_m \in W_r)$ is an indicator function that evaluates to $1$ if word $w_m$ is present in the recipe $r$ and evaluates to $0$ otherwise.
\subsubsection*{\textbf{Generative Modeling Setting}}\label{subsubsec:GM}
Consider the same definition of a set of recipes $R$ as shown in Eq \eqref{eqn:recipes}. Let $X_r = \{w_1,\hdots,w_L\}$ denote a partially complete natural language description of the recipe $r$ consisting of $L$ words, i.e., $X_r \subset W_r$
Formally, for a recipe $r \in R$, given a partial description $X_r$, the task is to learn a function $f: X_r \rightarrow g_r$ that can generate the set, 
\begin{equation}\label{eqn:gen}
    g_r = X_r \cup W_r\setminus X_r \cup A_r
\end{equation}
\subsection{Training and Evaluation Metrics}\label{subsec:metrics}
The training metrics we consider in our experiments are cross-entropy (CE) loss and weighted cross-entropy (WCE) loss. We use the WCE loss as there is a class imbalance issue, i.e., the number of cooking action words across all recipes compared to other words is relatively low. In the discriminative modeling case, the CE and WCE loss reduces to the binary CE and WCE losses, respectively. In the generative modeling case, only the CE loss (during word prediction of the words in the set defined in Eq \eqref{eqn:gen}) is considered. Since there is a class imbalance in the dataset, we do not use accuracy and instead use precision and recall as our evaluation metrics.

\subsubsection*{\textbf{Time and Space Efficiency Considerations}} \label{subsubsec:eff}
Despite the remarkable progress made by state-of-the-art AI models in recent years, they are not easily available to the general public. A key reason for this is their significant resource requirements, such as long training and inference times and high memory usage. For an AI system that can assist any individual in making healthy food choices, it is necessary for it to be resource-efficient, such as being able to run on a mobile phone and handle daily queries without overburdening the system. Consequently, we also evaluate the number of parameters and memory usage of the models used for experimentation in both discriminative and generative modeling settings.
\section{Methods and Evaluation}
We divide our methods into two categories, (A) methods that work in the discriminative modeling setting and (B) methods that work in the generative modeling setting. For each category of methods, we provide describe the methods, detail the experimental setup, and discuss their evaluation metric scores.

\subsection{Discriminative Modeling Methods and Evaluation}
\subsubsection*{\textbf{Data Preprocessing}} Given the discriminative modeling task description in Section \ref{subsubsec:DM}, our aim is to learn the binary function $f: S \rightarrow \{0,1\}$ that decides if a recipe word $w_m \in W_r$ is a cooking action or not. Since our dataset annotations consist of the full description of the cooking actions, e.g., for a recipe consisting of ''marinate the meat and leave overnight'', the annotation would be ``marinate'' (see Section \ref{subsec:data}). We manually preprocess the annotations to reflect binary ground truth labels corresponding to if a word $w_m \in W_r$ is a cooking action or not. For example, the annotation ``marinate'' is preprocessed as a one-hot-encoded vector of size $M$ for all the $M$ words in $S$, with $1$ in the places corresponding to the words ``marinate'' and $0$ for every other word.

\subsubsection*{\textbf{Method Descriptions}}
We try a rule-based method (R), where we use a simple lookup from a predefined list of cooking actions for each word  to label the word as a cooking action or not (the predefined list can be found \href{https://github.com/revathyramanan/cooking-action-generation}{here}). Next, we use the fine-tuned language models ELECTRA and XLNET \cite{clark2020electra,yang2019xlnet}. As mentioned in Section \ref{subsec:metrics}, the discriminative modeling task suffers a class imbalance issue. The models XLNET and ELECTRA have specifically been trained to be robust to class imbalances in the data. Finally, we also try industry standards that have shown particular success in ``binary tag prediction'' due to the discriminative modeling task essentially boiling down to predicting the tag of a word as either a cooking action or not a cooking action. We experiment with Stanford's NLP tools and explosion AI's SpaCy \cite{choi2012guidelines,finkel2005incorporating}.
\subsubsection*{\textbf{Experiments and Evaluation}}
\paragraph*{Model Configurations for the language models} For ELECTRA, we fine-tuned the electra-base-discriminator model obtained from hugging face. Early stopping was employed to prevent the model from overfitting. For XLNET, we fine-tuned the xlnet-large-cased model from hugging face. Similar to ELECTRA, early stopping was also implemented during training to prevent overfitting. For both ELECTRA and XLNET, the default pytorch implementations for CE and WCE loss were used. For the WCE loss, the weight assignment was $0.881$ for label $1$ and $0.119$ for label $0$.  
\paragraph*{Specifics of the SpaCy and Stanford's NLP model} We use Stanford's NER with its default configuration with the recognized named entities being labeled with $1$ (other words $0$), and SpaCy's part-of-speech (POS) tagger model `en-core-web-lg' to label cooking actions (label $1$ if POS is a ``verb'' for a word and $0$ for other words). 

\pgfplotstableread[col sep=comma,]{data.csv}\datatable
\begin{figure}[!htb]
    \centering
    \begin{tikzpicture}
\begin{axis}[legend style={at={(0.5,-0.25)},
		anchor=north,legend columns=-1}, anchor=north, width=\linewidth,
    ybar,
    bar width=3.5pt,
    xlabel={},
    ylabel={Discriminative Modeling Evaluations},
    xtick=data,
    xticklabel style = {rotate=30,font=\small},
    xticklabels from table={\datatable}{A},
    ymajorgrids,
    legend image code/.code={%
                    \draw[#1, draw=none] (0cm,-0.1cm) rectangle (0.2cm,0.1cm);
                }
             ]
    \addplot table [x expr=\coordindex, y= B]{\datatable};
    \addplot table [x expr=\coordindex, y= C]{\datatable};
    \legend{Precision, Recall}
\end{axis}
\end{tikzpicture}
    \caption{Performance results of various discriminative modeling methods to predict cooking actions. X-CE, E-CE, X-WCE, E-WCE, R, SpaCy, Stanford represents XLNET with CE loss, ELECTRA with CE loss, XLNET with WCE loss, ELECTRA with WCE loss, The rule-based system (R), SpaCy's POS tagger, and Stanford's NER model respectively.}
    \label{fig:dm_eval}
\end{figure}
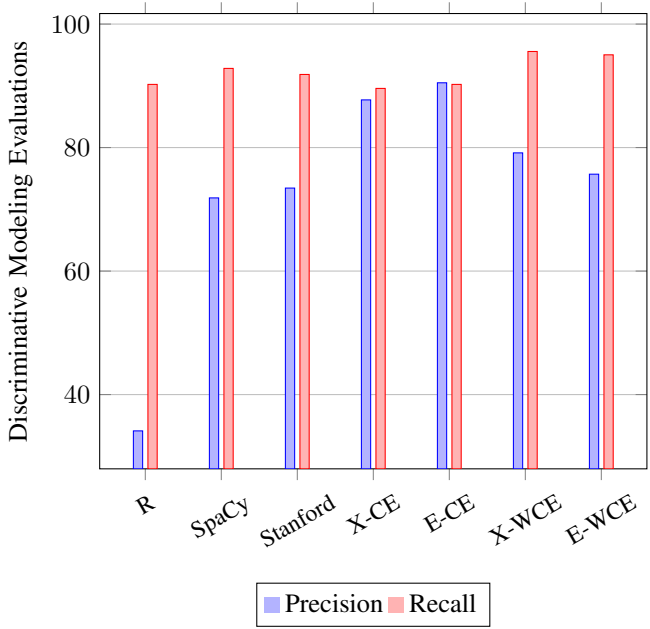

The evaluation scores for various discriminative modeling methods are presented in Figure \ref{fig:dm_eval}. The rule-based system (R) exhibits the worst performance, which is unsurprising given the challenges of creating precise rules for determining cooking actions from recipes, particularly when the data is characterized by irregular patterns. Although SpaCy's POS tagger and Stanford's NER model, the two industry standards, perform somewhat better, they employ Convolutional Neural Network and Conditional Random Field-based approximation models that can only capture linear chain structured contextual dependencies between the recipe words. As a result, we anticipate that self-attention transformer-based techniques such as XLNET and ELECTRA will outperform these models since they can model more complex interactions or dependencies between the recipe words and are also designed to be robust to irregular data patterns. As predicted, XLNET and ELECTRA demonstrate significant performance improvements. Nonetheless, employing WCE instead of CE in the training loss results in decreased precision, indicating that the model with CE loss may have overfit spurious data patterns. We conclude from these experiments that discriminative modeling may not be suitable for cooking action prediction since discriminatory patterns in the data are not present with regularity. Generative modeling has been demonstrated to be highly effective in situations where easily recognizable discriminatory patterns are absent, with the most notable example being large self-attention transformer-based models for language modeling. As a result, we will now move on to an experimental analysis in the generative modeling context.
\subsection{Generative Modeling Methods and Evaluation}
\subsubsection*{\textbf{Data Preprocessing}} Given the generative modeling task description in Section \ref{subsubsec:GM}, our aim is to learn the function $f: X_r \rightarrow g_r$, that generates $g_r$ as defined in Eq \eqref{eqn:gen}. Therefore we first tokenize recipe $r$ into a set of words $W_r$ and randomly sample the subset $X_r$ from $W_r$. The set of cooking actions $A_r$ is already part of the annotated dataset.
\subsubsection*{\textbf{Method Descriptions}}\label{subsubsec:cgen} 
While self-attention transformer-based generative models have shown impressive success in various tasks, their high resource usage makes them challenging for ordinary users. Additionally, scaling up the model parameters and data does not necessarily solve the problem of irregular data patterns. To address this issue, aggregation-based methods can be utilized to enhance the robustness of the model to such irregularities. These methods handle data pattern irregularities by learning ``aggregate patterns'' that regularly appear in the data. To create small-scale generative models that can also handle irregular data patterns, we conducted experiments using two methods that have proven universal approximation capabilities: (1) a two-layer feedforward neural network with ReLU activations that aggregates the layer outputs across input words, and (2) a polynomial approximation-based aggregation method that raises word representations to higher-order polynomial powers before aggregating them. We also incorporated a position embedding layer to provide information about the sequence of words in the recipes. In general, the proposed generative method \texttt{Cook-Gen} employs aggregation operations in a generative modeling setup and allows toggling between different generative modeling architectures, e.g., neural network-based approximators vs. polynomial powers-based approximators. We will now formally describe the forward pass of the implemented generative models that can be used with \texttt{Cook-Gen} to generate the cooking actions for $A_r$ for a recipe $r$ from a partial natural language description (set of words) of the recipe $X_r$.
\paragraph*{\texttt{Cook-Gen}-NN using an Aggregated Two-Layer Feedforward Neural Network Approximation}\label{par:nn}   
Using the same notation as in Section \ref{subsubsec:GM}, denoting $X_r$ as the input to the forward pass, we can describe the sequence of forward pass steps as follows:
\begin{align}
&E_r = \mathbf{E}X_r,~\mathbf{E} \in \mathbb{R}^{M \times d} \\
&H_r = E_r + \mathbf{P_e}X_r,~\mathbf{P_e} \in \mathbb{R}^{L \times d} \\
\begin{split}\label{eqn:nn}
&H_r = \max(\mathbf{W_z}^TH_r,0),~\mathbf{W_{1}} \in \mathbb{R}^{L \times d_{1}},~\mathbf{W_{z}} \in \mathbb{R}^{d_1 \times d_{2}},\\
&~z \in \{1,2\}
\end{split} \\
\begin{split}\label{eqn:aggr_1}
&H_r = \mathbb{E}(H_r[i,]),~i \in \{1,2,\hdots d_1\}
\end{split} \\
&O_r = \mathbf{W_o}^TH_r + \mathbf{B_o},~\mathbf{W_o} \in \mathbb{R}^{d_{2} \times M},~\mathbf{B_o} \in \mathbb{R}^{d_{2}\times M}
\end{align}
The final logits (before applying softmax) are obtained by extracting the last column of $M$ values from $O_r$. Eq \eqref{eqn:aggr_1} in the forward pass denotes the aggregation operation, i.e., taking the average (expected value) of all $d_1$ rows of the 2nd hidden layer output matrix. Here $d$ represents the embedding dimension, $\mathbf{E}$ denotes the embedding matrix, $\mathbf{P_e}$ denotes the position embedding matrix, and the $d_z$ corresponds to the dimensions of the hidden layer weight matrices.
\paragraph*{\texttt{Cook-Gen}-PF using an Aggregated Polynomials-based Function Approximation}\label{par:avg}
Neural networks are popular because the steps to obtain the hidden layer output matrix before the application of the aggregation operation in Eq \eqref{eqn:aggr_1} can be thought of as obtaining a feature map of the inputs $X_r$ using the neural network. Such a feature map is \textit{adaptive} because it is not predefined. Also, functions of feature maps obtained using neural networks, as described in Section \ref{par:nn} (two layers with ReLU activations), have universal approximation capabilities. However, we also know that a set of polynomial powers of the input $X_r$ is also a feature map, functions of which can also have universal approximation capabilities. But we first need to provide the function approximator knowledge of embeddings and positions of words in $X_r$. Thus we compute aggregates over polynomial powers of the rows in matrix $H_r$ obtained just before Eq \eqref{eqn:nn}, i.e., $H_r$ obtained after applying the embedding and position embedding matrix operations to $X_r$. The aggregation of polynomial powers of the rows of matrix $H_r$ is described as follows:
\begin{align}
&H_r = \{\mathbb{E}(H_r[i,]^j),~i \in \{1,2,\hdots L\}\mid j \in \{1,2,\hdots,J\}\}
\end{align}
We then get a set of $J$ vectors that are compiled into a new matrix $H_r$ of dimension $\mathbb{R}^{J\times d}$. We also need to correspondingly modify the dimensions of weights and biases $\mathbf{W_o}$ and $\mathbf{B_o}$ of the output layer to be of dimension $\mathbb{R}^{d\times M}$.
\paragraph*{\texttt{Cook-Gen}-GPT using ChatGPT}\label{par:gpt} Given that ChatGPT is a generative model, we aim to incorporate it into the suggested "Cook-Gen" approach. Despite ChatGPT requiring significant computational resources, it can be made accessible to the general public through API calls, which eliminates the need to store the model on their personal devices. Additionally, by leveraging backend infrastructure, the API calls can reduce inference times and latency. Figure \ref{fig:ChatGPT} illustrates how ChatGPT can be utilized to generate cooking instructions through API calls.
\begin{figure}[!htb]
    \centering
    \includegraphics[width=\linewidth, trim = 0cm 6.5cm 0cm 6.5cm, clip]{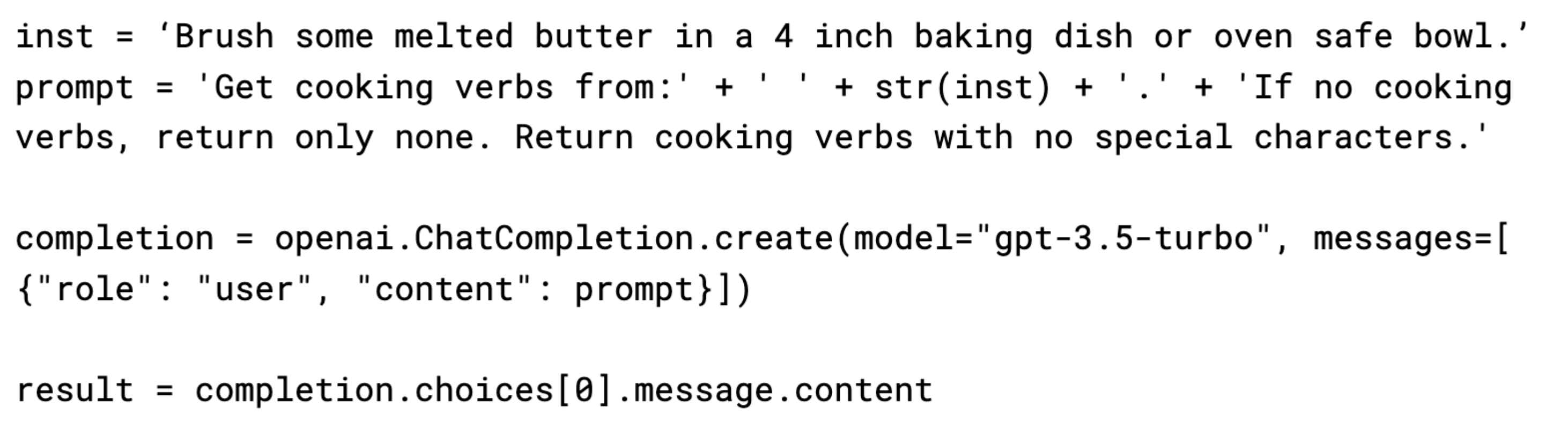}
    \caption{The figure shows how ChatGPT is prompted to obtain cooking actions from recipes using OpenAI API calls.}
    \label{fig:ChatGPT}
\end{figure}
\subsubsection*{\textbf{Experiments and Evaluation}}
\paragraph*{Model Configurations for \texttt{Cook-Gen}-NN, and \texttt{Cook-Gen}-PF}
The \texttt{Cook-Gen}-NN model is trained using a configuration with $d=d_z=200$ for $z$ values of 1 and 2, $L=512$, and $M=33000$. This results in a total of 13.38 million parameters. On the other hand, the \texttt{Cook-Gen}-PF model is trained with $d=200$, $L=512$, $J=3$, and $M=33000$, resulting in a total of 13.3 million parameters. Both models have trainable matrices consisting of $\mathbf{E}$, $\mathbf{P_e}$, ${\mathbf{W_z}}$ for $z \in {1,2}$, $\mathbf{W_o}$, and $\mathbf{B_o}$. Unlike the \texttt{Cook-Gen}-NN model, the number of parameters for the \texttt{Cook-Gen}-PF model does not depend on the polynomial order $J$. This means that the \texttt{Cook-Gen}-PF model has an advantage over the \texttt{Cook-Gen}-NN model in that it can handle more complex functions without increasing the parameter count.

\pgfplotstableread[col sep=comma,]{data2.csv}\datatables
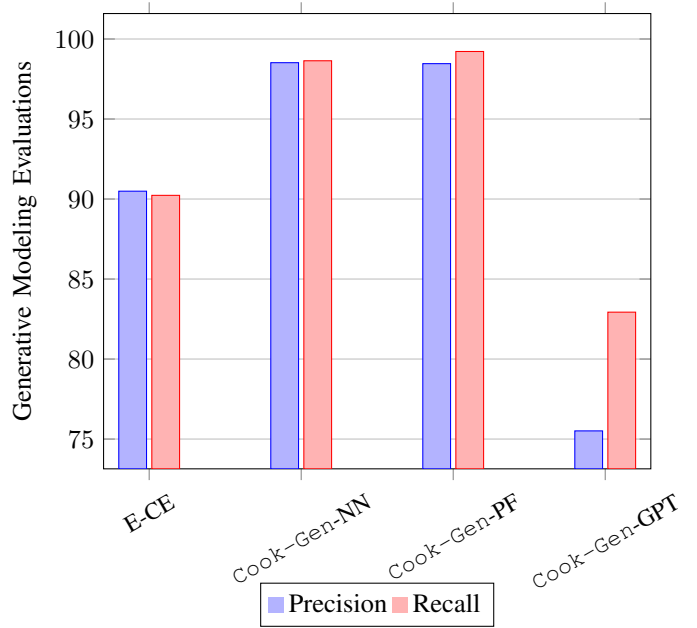
\begin{figure}[!htb]
    \centering
    \begin{tikzpicture}
\begin{axis}[legend style={at={(0.5,-0.25)},
		anchor=north,legend columns=-1}, anchor=north, width=\linewidth,
    ybar,
    bar width=10.5pt,
    xlabel={},
    ylabel={Generative Modeling Evaluations},
    xtick=data,
    xticklabel style = {rotate=30,font=\small},
    xticklabels from table={\datatables}{A},
    ymajorgrids,
    legend image code/.code={%
                    \draw[#1, draw=none] (0cm,-0.1cm) rectangle (0.2cm,0.1cm);
                }
             ]
    \addplot table [x expr=\coordindex, y= B]{\datatables};
    \addplot table [x expr=\coordindex, y= C]{\datatables};
    \legend{Precision, Recall}
\end{axis}
\end{tikzpicture}
    \caption{Performance results of various generative modeling methods used to generate cooking actions as part of \texttt{Cook-Gen} compared to the best performing discriminative modeling method E-CE (ELECTRA with CE loss). \texttt{Cook-Gen}-NN, \texttt{Cook-Gen}-PF, \texttt{Cook-Gen}-GPT, represents \texttt{Cook-Gen} with a two-layer neural network approximation, with a polynomial powers approximation, and with access to ChatGPT API calls, respectively.}
    \label{fig:gm_eval}
\end{figure}

The evaluation scores for the different variants of \texttt{Cook-Gen} is shown in Figure \ref{fig:gm_eval}. The figure also shows the best-performing discriminative modeling method for comparison. We see that \texttt{Cook-Gen}-PF, in which the polynomial approximation is used with aggregation, performs slightly better than \texttt{Cook-Gen}-NN, in which the two-layer neural network approximation is used. Although, the difference is not a lot. ChatGPT performs rather poorly. However, it is worth noting that we do not have access to the model and are unable to fine-tune it on our dataset. Fine-tuning ChatGPT could have resulted in improved performance. Thus, we see that generative modeling using our proposed aggregation methods significantly outperforms strong baselines, demonstrating the effectiveness of \texttt{Cook-Gen} for modeling cooking actions despite issues and challenges caused by irregular data patterns.

\pgfplotstableread[col sep=comma,]{data3.csv}\datatabless
\begin{figure}[!htb]
    \centering
    \begin{tikzpicture}
\begin{axis}[legend style={at={(0.5,-0.25)},
		anchor=north,legend columns=-1}, anchor=north, width=\linewidth,
    ybar,
    bar width=10.5pt,
    xlabel={},
    ylabel={Model Resource Consumption},
    xtick=data,
    xticklabel style = {rotate=30,font=\small},
    xticklabels from table={\datatabless}{A},
    ymajorgrids,
    legend image code/.code={%
                    \draw[#1, draw=none] (0cm,-0.1cm) rectangle (0.2cm,0.1cm);
                }
             ]
    \addplot table [x expr=\coordindex, y= B]{\datatabless};
    \addplot table [x expr=\coordindex, y= C]{\datatabless};
    \legend{GPU-Usage in GB, Params in Millions}
\end{axis}
\end{tikzpicture}
    \caption{The graph displays the usage of GPU memory (in GBs) and the number of parameters (in millions) for the ELECTRA and XLNET discriminative model methods in comparison to the \texttt{Cook-Gen}-NN and \texttt{Cook-Gen}-PF models. The resource consumption of the \texttt{Cook-Gen} models is considerably lower, indicating that \texttt{Cook-Gen} is highly effective for low-resource cooking action modeling.}
    \label{fig:res_eval}
\end{figure}
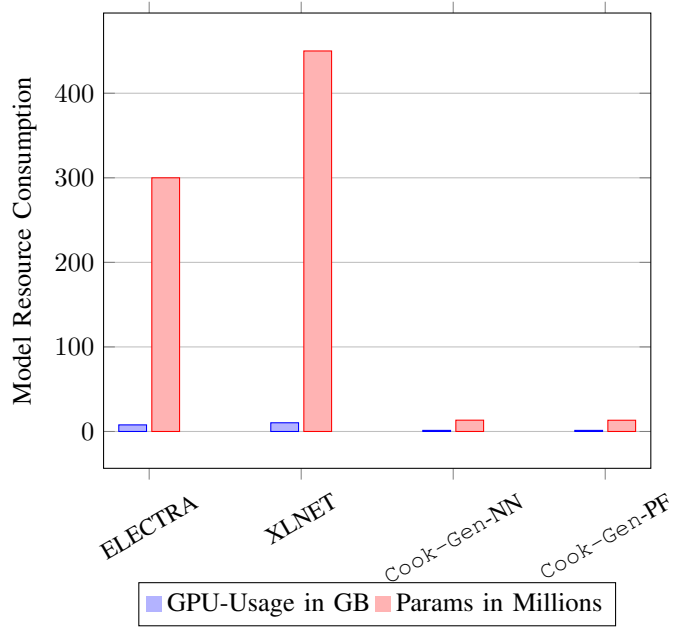
\subsection{Time and Space Efficiency Considerations}
In Section \ref{subsubsec:eff}, we discussed the need for efficient modeling of cooking actions to allow easy access for ordinary users that do not have access to the resources required to run large state-of-the-art models such as ChatGPT.

Figure \ref{fig:res_eval} shows the resource consumption in terms of GPU memory usage (in GBs) and the number of parameters in millions for the discriminative model methods ELECTRA and XLNET compared to the \texttt{Cook-Gen}-NN and \texttt{Cook-Gen}-PF models. The \texttt{Cook-Gen} models consume significantly fewer resources showing the effectiveness of \texttt{Cook-Gen} for low-resource cooking action modeling that is accessible to ordinary users on resource-constrained devices (e.g., mobile phones).
\section{Conclusion}
In this paper, we introduce \texttt{Cook-Gen}, which performs generative modeling of cooking actions that has the following salient features (1) \texttt{Cook-Gen} is robust to irregular data patterns in recipes caused by variations in cooking action descriptions and imbalanced word frequencies (2) \texttt{Cook-Gen} uses significantly lower memory-usage and parameters for modeling while still outperforming much larger baselines. \texttt{Cook-Gen} can be immensely useful as part of a large automated AI system that assists people with maintaining healthy eating habits, such as the recommendation system shown in Figure \ref{fig:cooking_process_aware_AI}.
\paragraph*{\textbf{Code and Data}} All the code and data used to produce the results in the paper will be made available at this \href{https://github.com/revathyramanan/cooking-action-generation}{link}
\section*{Acknowledgements}
This work was supported in part by the National Science Foundation under Grant 2133842, “EAGER: Advancing Neuro-symbolic AI with Deep Knowledge-infused Learning" and was carried out under the advisement of Prof. Amit Sheth \cite{sheth2023neurosymbolic,sheth2021knowledge,sheth2022process}. 
\bibliographystyle{IEEEtran}
\bibliography{references.bib}
\end{document}